\documentclass{ifacconf}

\usepackage{graphicx}       
\usepackage{color}          
\usepackage[utf8]{inputenc} 
\usepackage{amsmath}
\graphicspath{{figures/}}
\usepackage{natbib}         
\usepackage{balance}

\usepackage{floatrow}
\usepackage{subfig}

\begin{document}
\begin{frontmatter}

\title{Dense Disparity Estimation in Ego-motion Reduced Search Space\thanksref{footnoteinfo}} 

\thanks[footnoteinfo]{This work has been supported by the European Union’s Horizon~2020 research and innovation
programme under grant agreement No. 688117 and by the Croatian Science Foundation under contract No. I-2406-2016. This
research has also been carried out within the activities of the Centre of Research Excellence for Data Science and
Cooperative Systems supported by the Ministry of Science, Education and Sports of the Republic of Croatia.}

\author[First]{Luka Fućek, Ivan Marković, Igor Cvišić, Ivan Petrović} 

\address[First]{University of Zagreb, Faculty of Electrical Engineering and Computing, 
   Croatia (e-mail: luka.fucek@fer.hr, ivan.markovic@fer.hr, igor.cvisic@fer.hr, ivan.petrovic@fer.hr).}

\begin{abstract}                
Depth estimation from stereo images remains a challenge even though studied for decades.
The KITTI benchmark shows that the state-of-the-art solutions offer accurate depth estimation, but are still 
computationally complex and often require a GPU or FPGA implementation.
In this paper we aim at increasing the accuracy of depth map estimation and reducing the computational complexity by
using information from previous frames.
We propose to transform the disparity map of the previous frame into the current frame, relying on the estimated
ego-motion, and use this map as the prediction for the Kalman filter in the disparity space. 
Then, we update the predicted disparity map using the newly matched one.
This way we reduce disparity search space and flickering between consecutive frames, thus increasing the
computational efficiency of the algorithm.
In the end, we validate the proposed approach on real-world data from the KITTI benchmark suite and show that the
proposed algorithm yields more accurate results, while at the same time reducing the disparity search space.
\end{abstract}

\begin{keyword}
stereo vision, visual odometry, disparity estimation.
\end{keyword}

\end{frontmatter}

\section{Introduction}\label{sec:intro}

Depth estimation plays an important role in many autonomous systems in automotive industry, augmented reality, and robotics
in general.
Besides time-of-flight cameras, which can be limited by external lighting and range, stereo cameras are often used as a
primary vision sensor for depth estimation. 
Even though using cameras requires significant computing efforts to obtain disparity maps (DM), 
stereo cameras are a very popular choice due to their commercial availability and efficiency.

In order to obtain valuable depth information, images from calibrated and synchronized stereo cameras are used.
Early research was focused on local methods that considered only neighboring pixels to find a stereo match.
Lack of rich texture in the stereo images resulted with semi-dense DM with poor accuracy.
To tackle these problems, pixel-wise global methods penalizing discontinuities were introduced, where a depth map is
sought minimizing a global energy function, e.g., works of \cite{Kolmogorov2001,Klaus2006,Yang2009}.
However, global methods incur high computational and memory costs and in \cite{Hirschmuller2008} semi-global
matching (SGM) algorithm was proposed.
Therein the computational complexity of global optimization methods was reduced by reducing the domain of considered
pixels to several linear paths in the images.
Even though the DM obtained by SGM is dense and more accurate than the ones obtained with 
local matching techniques, they still lack temporal information and can introduce depth flickering among neighboring
frames.

As opposed to standard approaches that extract depth information from a single stereo image,
\cite{Dobias2011} predict the DM using ego-motion and DM of the previous frame.
They take predicted disparities as they are and fill the rest of the DM using a traditional stereo algorithm,
while checking the validity of predicted disparities without any method to refine them.
%
%
Additionally, \cite{Jiang2014} detect moving objects in the scene and avoid predictions based on ego-motion in these regions of the
image.
However, their algorithm accumulates propagated disparity error with time, since they also take predicted
disparities without any refinement.
To address this problem, the authors discard the predicted DM every 100 frames to start from a fresh
SGM-based DM. 
From the viewpoint of accuracy, theoretically this result does not outperform SGM, since unreasonable disparities are
propagated into future frames until the reset is performed.
From estimation perspective, \cite{Vaudrey2008} integrate previous frames with the current one using an iconic (pixel-wise) Kalman filter 
as introduced by \cite{Matthies1989}.
Their work extends the idea of integrating stereo iconically to provide more information with a higher certainty.
They extend the Kalman filter model by introducing \textit{disparity rate} in the depth direction. 
However, they constrain their model to motion in longitudinal direction and, thereby, neglect all movements in lateral and vertical directions.
Their algorithm performs well only in scenes where most of the movement is in the longitudinal direction, such as highway traffic scenes.
\cite{Morales2013} aim to improve the disparity estimation of objects that are static with respect to the ground or moving longitudinally away from the ego-vehicle.
However, they also constrain their motion model (cf. \cite{Franke2005}). 
\cite{JakobEngelJorgStuckler2015} consider ego-motion and form keyframes 
that contain depth information for simultaneous localization and mapping (SLAM). 
They integrate the disparity information from current frame into the assigned keyframe.
Since the focus of their work was SLAM, DM generated by their approach are semidense.
\cite{Zbontar2015} use convolutional neural networks (CNN) to learn a similarity measure on small image patches and 
compute matching costs. SGM is then used to optimize the results. 
\cite{Mayer2015} formulated the problem as a supervised learning task that can be solved with CNN.
They proposed three synthetic stereo video datasets for training of large networks
and presented a CNN for real-time disparity estimation that provides state-of-the-art results
using high power graphics processing units.

In this paper we propose to generalize the disparity prediction model and increase the accuracy of depth estimation
in comparison to standard methods like the SGM.
We focus our work on stable, precise and fast spatio-temporal reconstruction, thus constraining the usecase of the
proposed method to static scenes.
Although this can be seen as a limitation, in fact, this approach will form the base for dense stereo detection of
dynamic objects and can also be used in applications where static scenes are predominant, e.g., space exploration
robotics.
We track pixels from the previous stereo frame to the current one directly in the disparity domain using ego-motion
estimation. 
In order to avoid the need to introduce any other sensors, we obtain ego-motion using the visual odometry algorithm
proposed by \cite{Cvisic2015}.
For each pixel we deterministically compute the displacement based on ego-motion and
stochastically track the value of its disparity while updating its uncertainty through time with Kalman filtering.
Disparity of each pixel is estimated by combining the newly matched (measured) DM and predicted DM.
We perform stereo matching using a custom SGM on a reduced disparity search space, based on the predicted DM and
its uncertainty to reduce the computing complexity and number of matched outliers, while producing denser and more
accurate DM.
In the end, we validate the proposed approach on real-world data from the KITTI dataset (\cite{Menze2015CVPR})

\section{Algorithm description}

As an input for stereo disparity estimation, we use a sequence of stereo image pairs, captured using a pair of
calibrated, rectified, and synchronized cameras.
The proposed algorithm relies on using the previous and current stereo image, $I^{k-1}$ and $I^{k}$, with the
accompanying DM of the previous frame, $D^{k-1}$, in order to estimate the DM of the current frame
$D^{k}$.
Each stereo frame at time instant $k$ consists of a left and right image, $I^k_L$ and $I^k_R$, and the accompanying
left and right DM, $D^k_L$ and $D^k_R$.
We use two consecutive stereo images, $I^{k-1}$ and $I^{k}$, for \textit{ego-motion estimation} of the stereo rig.

Given the estimated displacement calculated from $I^{k-1}$ and $I^{k}$, and the DM $D^{k-1}$ from the
previous frame, we can predict the DM of current frame $D^{k|k-1}$.
This \textit{disparity prediction} serves as the base for a \textit{stereo matching} technique, e.g., in the present
paper we use SGM, to reduce the disparity search space, thus consequently reducing the 
computing efforts and the number of outliers in the measured (matched) DM $D^k_{m}$.
To increase the accuracy and produce a denser DM, the predicted DM, $D^{k|k-1}$, is updated with
the measured DM, $D^k_{m}$, within the framework of the Kalman filter.
This step produces the updated disparity $D^{k|k}$ and we denote it as the \textit{disparity update}.
The algorithm can be summed up as follows:

\begin{enumerate}
  \item Ego-motion estimation
  \item Disparity prediction
  \item Stereo matching
  \item Disparity update.
\end{enumerate}


\subsection{Ego-motion estimation}

In order to predict current DM $D^{k|k-1}$ based on previous frame DM $D^{k-1}$, an
ego-motion estimation is needed.
Transformation $T_{k-1,L}^k$ represents the homogeneous transformation from the current coordinate frame of the left
camera, $\mathcal{F}^{k}_L$, to the previous frame $\mathcal{F}^{k-1}_L$.
This transformation can be obtained using several different methods.
In the present paper, we choose the visual odometry approach.

\subsubsection{Disparity space.}

We perform prediction directly in disparity space and, therefore, we transform $T_{k-1,L}^k$ from Euclidean space to
disparity space.
Let $w$, $M$ and $\Gamma$ be defined as 
\begin{equation} \label{eq:wMGdef}
  \omega =
  \begin{bmatrix}
    x \\
    y \\
    d \\
    1
  \end{bmatrix}
  ,
  M = 
  \begin{bmatrix}
    X \\
    Y \\
    Z \\
    1
  \end{bmatrix}
  ,
    \Gamma =
  \begin{bmatrix}
    f && 0 && 0 && 0 \\
    0 && f && 0 && 0 \\
    0 && 0 && 0 && fb \\
    0 && 0 && 1 && 0 \\
  \end{bmatrix},
\end{equation}
where $x, y$ and $d$ are coordinates in the disparity space, $X, Y$ and $Z$ are coordinates in the Euclidean space
camera frame $\mathcal{F}^{k}_L$, $b$ is the baseline of stereo rig and $f$ is the focal length of the cameras.
$\Gamma$ represents a projective transformation between homogeneous coordinates $M$ in Euclidean space and homogeneous coordinates  $\omega$ in disparity space (\cite{Demirdjian2001}).
Analogous to the standard homogeneous transformation in Euclidean space
\begin{equation} \label{eq:Mtrans}
  M^{k} =  T_{k-1}^k M^{k-1}
\end{equation}
the transformation of coordinates in disparity space is defined as
\begin{equation} \label{eq:wtrans}
  \omega^{k*} \simeq H_{k-1}^k \omega^{k-1}
\end{equation}
where $\simeq$ denotes equality up to a scale factor and $H$ is defined as
\begin{equation}
  H_{k-1}^k =  \Gamma T_{k-1}^k \Gamma^{-1}.
\end{equation}
%


\subsection{Disparity prediction}
\label{s:disp_pred}

Using the ego-motion $H_{k-1}^k$ we transform each pixel from $D^{k-1}$ to estimate $D^{k|k-1}$.
The transformation is performed directly in the disparity space using \eqref{eq:wtrans}. 
Note that in \eqref{eq:wtrans} the result of each pixel transformation is only equal up to a scale factor.
Since $\omega$ is expected to be homogeneous as in (\ref{eq:wMGdef}), we scale $\omega^{k*}$ by dividing it with the value of
its fourth member to get a homogeneous coordinate form $\omega^k$.
%
%

\subsubsection{Uncertainty prediction.}

While new $x$, $y$ and $d$ coordinates of each pixel are deterministically calculated by (\ref{eq:wtrans}), each
disparity $d^{k-1}$ in $D^{k-1}$ is associated with an accompanying variance $p_d^{k-1}$,
forming a variance map $P^{k-1}$ that includes left and right DM variances $P^{k-1}_L$ and $P^{k-1}_R$. 
Variance of each disparity is predicted using the motion model applying the same displacement
\begin{equation} \label{eq:DUMMY}
{p_d^{k|k-1}} = ({\Phi^{k-1}})^2 {p_d^{k-1}} + q^{k-1},
\end{equation}
where $q^{k-1}$ denotes the variance of estimated ego-motion in the disparity space,
and $\Phi^{k-1}$ denotes the system model of current transformation for a pertaining pixel.
The variance of estimated ego-motion depends on the precision of the used odometry algorithm.
How to compute odometry variance and transform it to disparity space is out of the scope of the present paper,
and in lieu of a time varying $q^{k-1}$, we use an empirically determined constant $q$.
Since we already know $d^{k|k-1}$, for variance estimation we compute the system model as the
following ratio
\begin{equation}
\Phi^{k-1} = \frac{d^{k|k-1}}{d^{k-1}}
\end{equation}
thus avoid the need for computing $\Phi^{k-1}$ analytically.

\subsubsection{Disparity refinement.}

When transforming each pixel from $D^{k-1}$, it is likely that multiple pixels from frame $k-1$ will result in the same
$x$ and $y$ coordinates in $D^{k|k-1}$.
If there are no outliers in the prediction process, pixels with the highest disparity value $d^{k|k-1}$ are the closest
ones to the camera and are most likely the pixels not being occluded by other pixels. 
Therefore, we select these pixels and discard the others.

Since we take a deterministic approach to $x$ and $y$ coordinates propagation described with \eqref{eq:wtrans},
incorrect disparity predictions are expected near object edges. 
Using ego-motion information $T_{k-1}^k$ and previous disparity $d^{k-1}$ we predict the observation of $d^{k|k-1}$.
In case of bad ego-motion estimation, prediction $d^{k|k-1}$ could result with wrong disparity value.
If the disparity is placed away from depth discontinuities, disparity prediction error will be small enough and \textit{stereo matching} in a close interval around this disparity
will easily correct the wrong prediction (see Section~\ref{s:ster_mtch}).
On the other hand, if we consider disparity $d^{k-1}$ that is placed on the depth discontinuity (edge),
prediction based on bad ego-motion could result with disparity error that can not be corrected in
\textit{stereo matching} phase.
To address this problem, we reject all the predicted pixels near disparity discontinuities.

%

Additionally, if the stereo rig is moving forward, gaps in the predicted DM appear.
This phenomenon can be compared to the \emph{zooming} effect when lack of information results in holes in the predicted
disparity (confer Fig.~\ref{fig:disp_proc_b} for an illustration of this effect).
We address this problem with task-specific interpolation that fills the \textit{invalid} disparities caused by this
phenomenon. 
We use horizontally neighboring pixels and use their values to determine the value of an \textit{invalid} disparity.
If $d_i$ is an \textit{invalid} predicted disparity in an image row, its value is determined as 
\begin{equation}
\begin{cases}
 d_i = 
    \frac{d_{i-1}+d_{i+1}}{2},    & \quad \text{if } f_s(d_{i-1}, d_{i+1}) = 1 \\
    \mathrm{invalid},  			& \quad \text{else}\\
  \end{cases}
\end{equation}
where $f_s$ represents a similarity function described as
\begin{equation}
f_s(a,b) = 
\begin{cases}
  1,    & \quad \text{if } |a - b| < \gamma_f \\
  0,    & \quad \text{else}\\
 \end{cases}
\end{equation}
where $\gamma_f$ denotes a threshold value defining if $a$ and $b$ (in our case, $d_{i-1}$ and $d_{i+1}$) are similar
enough.
In other words, we compare neighboring disparities and, if they are similar enough, we use their values to fill the
\textit{invalid} pixels.
We take the same approach for filling invalid values based on vertical neighboring pixels.

\subsection{Stereo matching} \label{s:ster_mtch}

Once the predicted DM, $D^{k|k-1}$, and its variance map, $P^{k|k-1}$, is available, we use this
information to additionally reduce the computational complexity of the process of stereo matching.
For the stereo matching we use SGM with eight-path configuration. 
For each path an energy function that penalizes disparity changes among neighboring pixels is minimized to find the
optimal disparity value for each pixel.
Unlike its usual implementation, we reduce the disparity search space based on the predicted DM $D^{k|k-1}$
and ${P^{k|k-1}}$.
Instead of searching the whole disparity space for every pixel at every iteration of stereo matching, we only consider
small environment centered around the predicted disparity $d^{k|k-1}$.
The reduced search interval is defined as $[d^{k|k-1}\pm 3\sqrt{p^{k|k-1}}]$.


\subsubsection{Matching uncertainty.} \label{s:mch_uncert}

To estimate the variance of each matched pixel, we refer to the approach described in \cite{wedel2011} 
where authors observe the slope from left and right side of the cost function around its minimum. 
The slope serves as a quality measure of the estimated disparity. 
If the slope is low, the disparity is not estimated precisely and the variance is higher. 
If the slope is high, the disparity is estimated precisely and the variance is lower.
\cite{wedel2011} have shown that uncertainty determined using this technique is correlated to the true
variance of the errors based on comparison with ground truth.
The disadvantage of this approach lies in the case when there are two or more neighboring pixels with minimal cost. 
This way the slope is horizontal and the uncertainty is infinite which often is not the case.
%

\begin{figure}[!t]
\centering
   \includegraphics[width=0.7\linewidth]{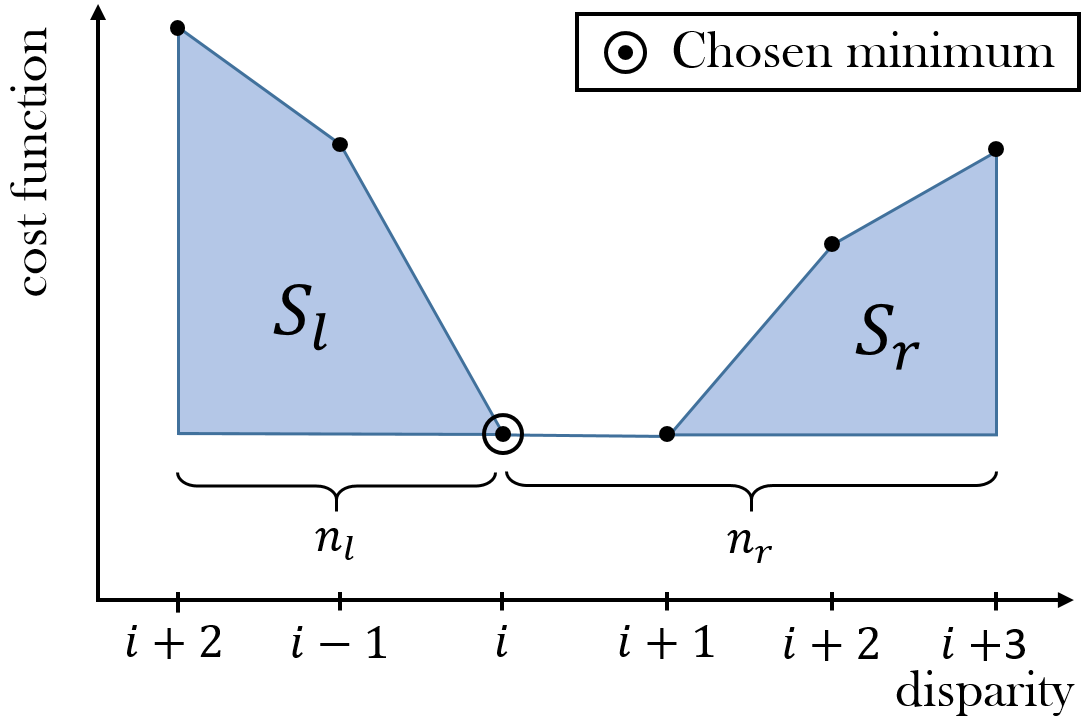}
   \caption{Variance estimation by counting neighboring disparities from left and right side of the cost function around its minimum.}
   \label{fig:var_b} 
\end{figure}

We propose to expand the variance estimation approach by counting neighboring disparities $n_l$ and $n_r$ while sums
$S_l$ and $S_r$ of their costs
are lower than $S_{max}$ (Fig. \ref{fig:var_b}). 
In our experiments, we determined that for matching window [3x3], $S_{max} = 10$ performs well over a wide range of scenes.
The matching variance $r^k$ is then defined as
\begin{equation}
r^k = (n_l+n_r).
\end{equation}
By considering multiple nearby matching cost, we avoid infinite uncertainties.
Additionally, the variance value does not depend on the immediately neighboring costs alone, but also on the costs that
fit into sum $S_{max}$.

Alternatively, innovation variance $s^k = p^{k|k-1} + r^k$ could be used for disparity search space reduction.
However, in order to determine $s^k$, we would need to first calculate the matching variance $r^k$.
Since $r^k$ depends on the disparity value and nearby optimization function costs, which are available only after the
matching process, we would need to rerun the matching process (on a wider interval), which would induce
additional computational load.

\subsubsection{Temporal stability.}

In \cite{Jiang2014}, disparities are propagated from previous moment $k-1$ to current moment $k$ as they are. 
If an unreasonable value (outlier) is present in the previous frame, it will also be propagated to the current frame. 
Since there is no mechanism to forget these unreasonable values, they will be propagated to every future frame and the
number of outliers in the final DM will increase with time.
\cite{Jiang2014} address this by forgetting the predicted disparity every 100 frames. 
In this work, we aim at continuous and constant improvement in accuracy and computing time, hence no predicted pixel is
taken as it is.
Predicted value is only used to reduce the disparity search space and its variance is tracked for further filtering as
described in Section~\ref{s:disp_updt}.

\subsection{Disparity update}\label{s:disp_updt}

Combining the predicted DM $D^{k|k-1}$ and the matched DM $D^k_{z}$, i.e. the measured disparity,
both with their respective variances, we update the predicted disparity using Kalman
filter on the pixel level. 
As described in Section \ref{s:disp_pred} we predict the new $x$ and $y$ coordinates of each disparity
deterministically, while disparity $d$ is estimated by Kalman filtering. 
First we determine the Kalman gain:
\begin{equation}
K_k = \frac{p_{k|k-1}}{p^{k|k-1}+r^k}
\end{equation}
where $r_k$ denotes the variance of the matched disparity (measurement) as described in Subsection \ref{s:mch_uncert}.
The final disparity is estimated using the standard Kalman filter update equation:
\begin{equation}
d^k = d^{k|k-1}+ K_k (d^{k}_z - d^{k|k-1})
\end{equation}
and the variance is updated as follows
\begin{equation}
p^{k} = (1-K_k) p^{k|k-1}.
\end{equation}


\subsubsection{Disparity consistency check.} \label{s:disp_filt}

Disparity consistency check is performed for two reasons.
First, outliers present in the current DM can be propagated to the next frame.
Second, even though propagated disparity is only used to reduce search space, the real minimum of the cost function
 could be outside of the reduced search space.
We address this problem by using several filtering techniques for rejecting outliers like \textit{left-right consistency
check} and \textit{sum of absolute difference (SAD) check} on left and right images.
Using this method we reject occluded pixels that are often poorly matched and which introduce outliers being propagated
to future frames.
In the next frame, for the rejected pixels, the search is then performed in the whole search space.
Please confer \cite{Hu2010a} for an overview of alternative methods.

\section{Experimental validation} \label{s:exp_val}

To test and validate the proposed algorithm we used the KITTI dataset (\cite{Menze2015CVPR}), which includes 200 training and testing
scenes. 
Every scene consist of 20 stereo sequences of real world images, thus making them appropriate for 
validation of algorithms that rely on image sequences rather than a single image frame. 
Since KITTI dataset is captured on the streets of Karlsruhe, moving objects are present in the majority scenes.
In order to evaluate the proposed algorithm, we selected scenes where no moving objects were present.
Current implementation treats the scene as static and any moving objects will introduce prediction errors since their movement is neglected, thereby resulting with bad final DM estimation.
As discussed in Section~\ref{sec:intro}, this limitation is in fact a base for future dense scene flow estimation and
moving object detection.
Furthermore, it can also be used in applications where moving objects are not dominant or even non-existent, like space
exploration robotics.

\begin{figure}[!t]
\centering

  \sidesubfloat[]{\includegraphics[width=0.9\linewidth]{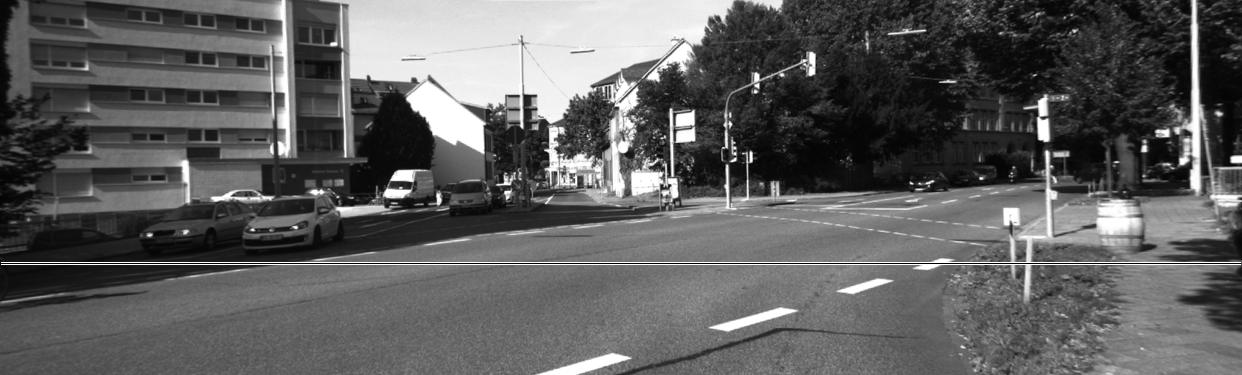}\label{fig:red_spac_a}}\quad%
  \sidesubfloat[]{\includegraphics[width=0.9\linewidth]{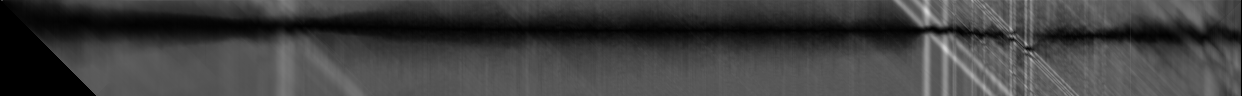}\label{fig:red_spac_b}}\quad%
  \sidesubfloat[]{\includegraphics[width=0.9\linewidth]{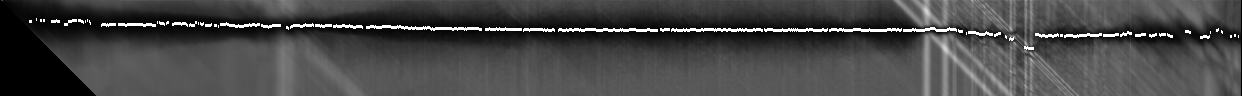}\label{fig:red_spac_c}}\quad%
  \sidesubfloat[]{\includegraphics[width=0.9\linewidth]{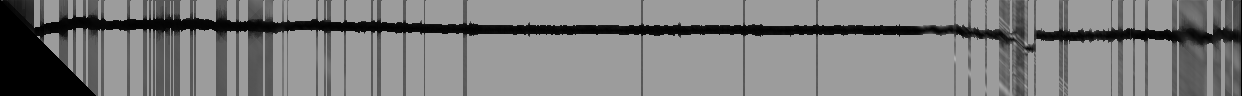}\label{fig:red_spac_d}}\quad%
  \sidesubfloat[]{\includegraphics[width=0.9\linewidth]{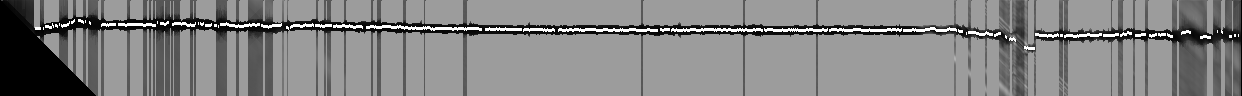}\label{fig:red_spac_e}}%

\caption
{Disparity search space reduction on an optimized matching cost function. 
(a) shows the left image of stereo frame with highlighted path example,
(b) shows optimized matching costs on the path,
(c) highlights the chosen minima (disparities)
(d) shows optimized matching costs with the reduced disparity search space
(e) highlights the chosen minima (disparities) after search space reduction.
}
\label{fig:red_spac}
\end{figure}

As described in Section~\ref{s:ster_mtch} we use SGM with disparity estimation in the reduced search space based on
ego-motion.
Demonstration of an arbitrary optimized path with matching costs is shown in Fig.~\ref{fig:red_spac}.
Figure~\ref{fig:red_spac_b} shows optimized matching costs of the path shown in Fig.~\ref{fig:red_spac_a} as a product
of our basic SGM implementation.
Next, Fig.~\ref{fig:red_spac_c} highlights the chosen minima, i.e., chosen disparity values on the whole path.
The idea of the proposed algorithm is to reduce the disparity search space based on valid disparity predictions as shown
in Fig.~\ref{fig:red_spac_d}.
Solid gray color represents the ignored disparity search space.
After the search space reduction, minima can be located faster while reducing the chance of false disparity detection.
Figure~\ref{fig:red_spac_e} highlights the reduced search space chosen minima, from which reduction of required
computation effort can be seen.

In the first iteration, whole disparity search space is considered since there is no disparity prediction available. 
In the following frames the disparity prediction is available and SGM can be performed on the reduced disparity search
space.
Our experiments show that less than 50\% of whole disparity search space is considered in most scenes. 
For the scene shown in Fig.~\ref{fig:red_spac}, the considered disparity search space is 100\%, 49.05\%, 47.24\% and
46.05\%
for frames 0, 1, 2 and 3, respectively. 
As the prediction becomes denser the whole computing process becomes faster due to the reduced search space.
%


\floatsetup[figure]{style=plain,subcapbesideposition=top}

\begin{figure}[!t] 
\centering

\sidesubfloat[]{\includegraphics[width=0.9\linewidth]{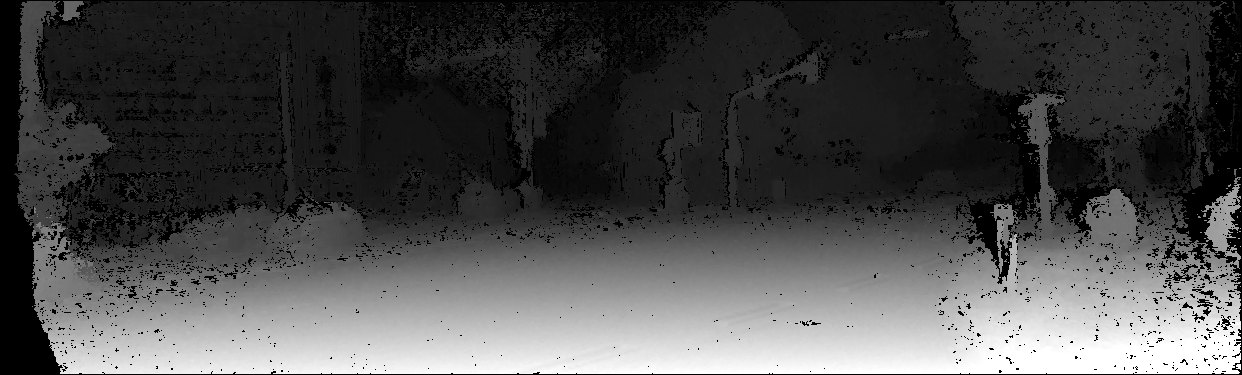}\label{fig:disp_proc_a}}\quad%
\sidesubfloat[]{\includegraphics[width=0.9\linewidth]{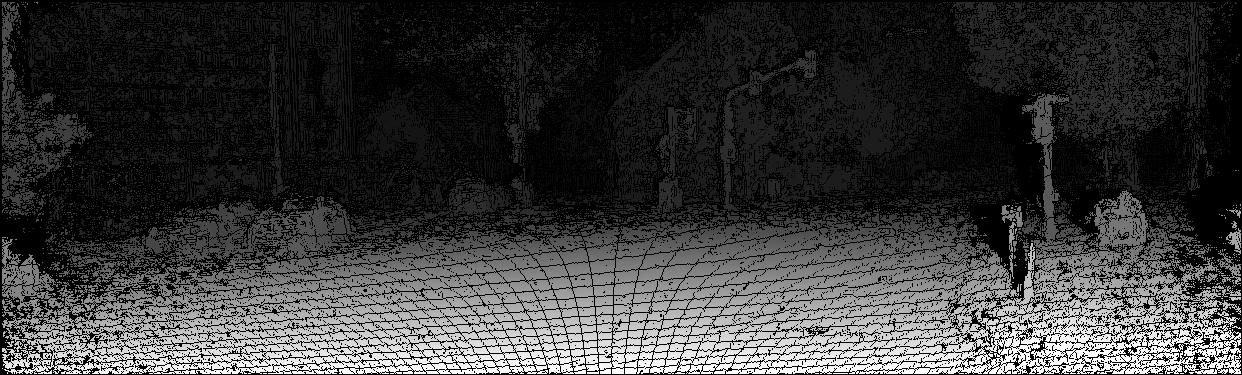}\label{fig:disp_proc_b}}\quad%
\sidesubfloat[]{\includegraphics[width=0.9\linewidth]{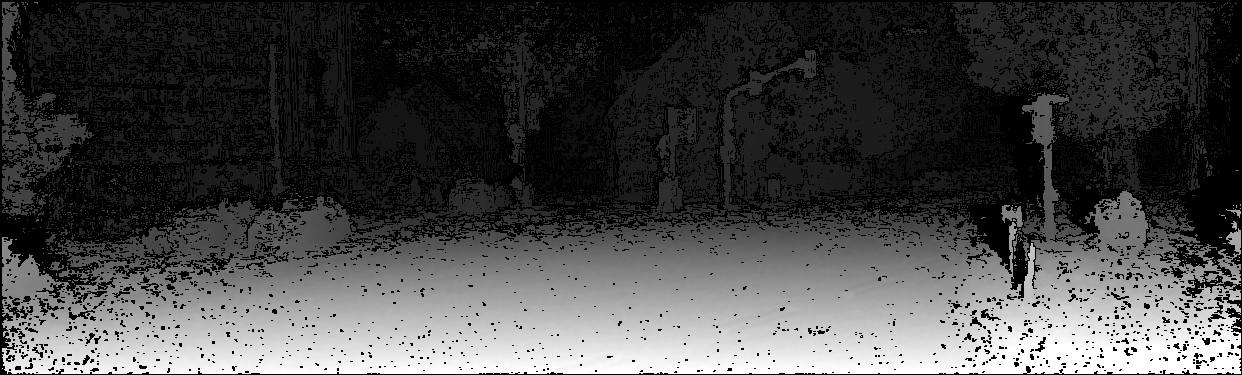}\label{fig:disp_proc_c}}\quad%
\sidesubfloat[]{\includegraphics[width=0.9\linewidth]{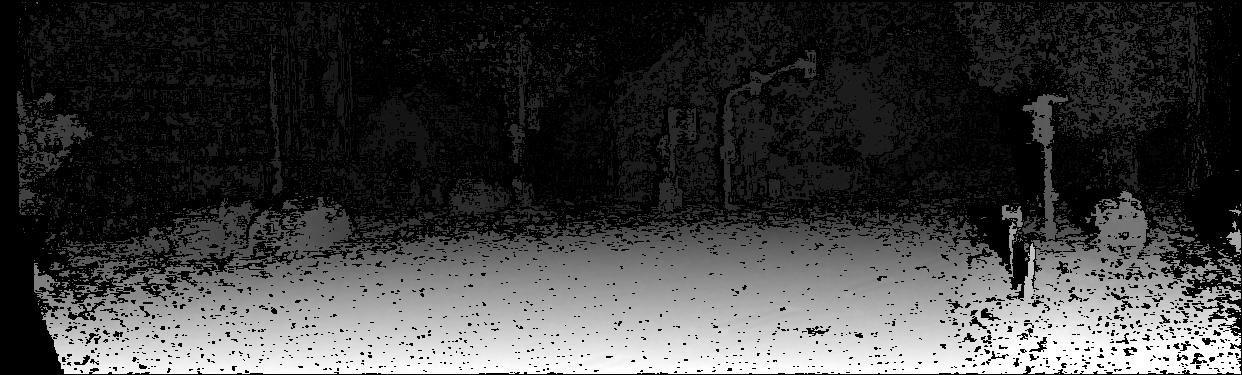}\label{fig:disp_proc_d}}\quad%
\sidesubfloat[]{\includegraphics[width=0.9\linewidth]{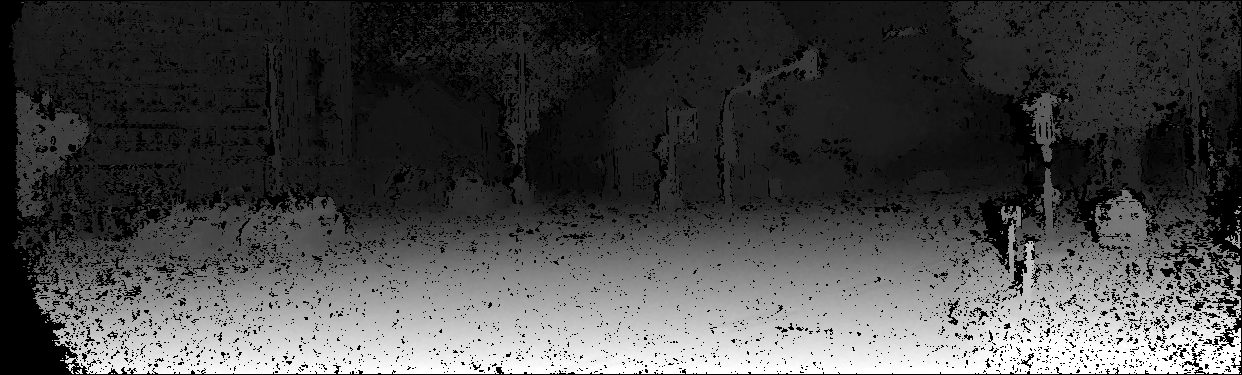}\label{fig:disp_proc_e}}\quad%
\sidesubfloat[]{\includegraphics[width=0.9\linewidth]{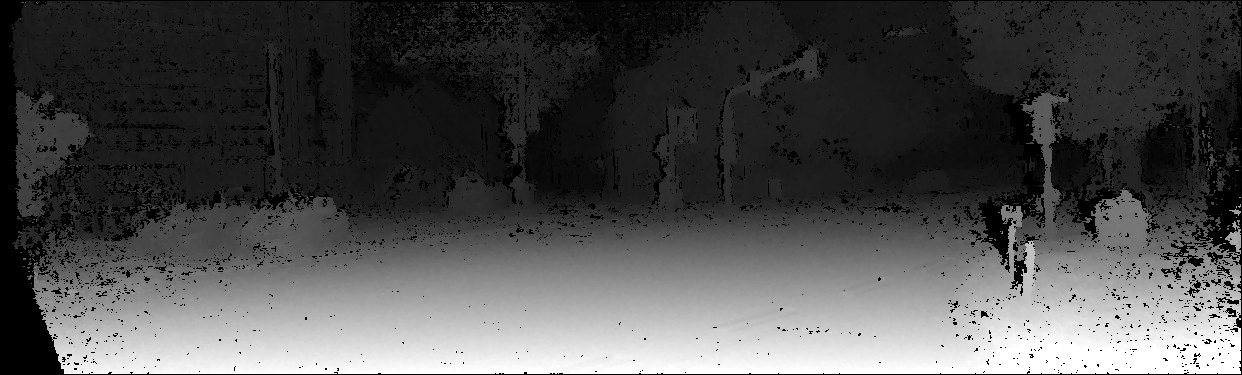}\label{fig:disp_proc_f}}\quad%

\caption
{Demonstration of disparity processing. Images show left DM of
(a) previous frame,
(b) prediction of current frame based on ego-motion and previous DM,
(c) prediction of current frame interpolated due to ``zooming'' effect,
(d) prediction of current frame filtered with left-right consistency check,
(e) SGM matching result on reduced search disparity space,
(f) updated (d) with (e).
}
\label{fig:disp_est}
\end{figure}

Figure \ref{fig:disp_est} demonstrates the whole disparity estimation process.
First we take the DM of the previous frame (Fig.~\ref{fig:disp_proc_a}) and form a prediction based on
ego-motion (Fig.~\ref{fig:disp_proc_b}).
Here, the ``zooming'' effect is easily noticeable.
We successfully negate that effect by interpolation method described in Section~\ref{s:disp_pred} as shown in
Fig.~\ref{fig:disp_proc_c}.
Next, left-right consistency check is performed on left and right DM to compare their results 
and discard all disparity predictions that are not below the similarity threshold.
This method is used to reduce a number of outliers caused by incorrect ego-motion propagation or poorly matched
disparities from previous frames.
The most noticeable rejected area is visible on the left side of Fig. \ref{fig:disp_proc_d} where all the
disparities
that cannot be observed with both left and right cameras are marked as invalid.
Fig.~\ref{fig:disp_proc_d} also represents the final DM prediction used for further processing.
Fig.~\ref{fig:disp_proc_e} shows the results of our implementation of an eight-path SGM algorithm with reduced disparity
search space. 
It is noticeable that predicted and matched DM (Figs.~\ref{fig:disp_proc_d} and \ref{fig:disp_proc_e}) are
very similar. 
The main differences are in the areas of occluded pixels due to different perspectives of previous and current stereo
frames.
Other noticeable difference is near object edges where edge rejection filtering described in Section~\ref{s:disp_pred}
is applied after the prediction step.
These rejections are necessary to reduce the number of outliers that could propagate to future frames 
and result with wrong disparity search interval in the matching process.
Finally, we update the DM as described in Section~\ref{s:disp_updt}.
The results are shown in Fig.~\ref{fig:disp_proc_f}.

\begin{figure}[!t] 
  \centering
    \includegraphics[width=0.65\textwidth]{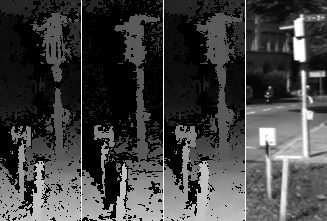} 
    \caption{The comparison of 
    (1) matched DM,
    (2) predicted DM,
    (3) updated DM and
    (4) intensity image.}
    \label{fig:compare}
\end{figure}

We additionally demonstrate the benefits of this approach by referring to Fig.~\ref{fig:compare}, 
where it is visible that the updated DM combines best disparities from both the matched and predicted
DM.
As elaborated in Section~\ref{s:disp_pred}, rejected disparities near disparity discontinuities 
can be seen in the predicted DM in Fig.~\ref{fig:compare}.
While useful for preserving the edges and shape of foreground objects, this rejection can result with holes 
in the predicted background disparities.
On the other hand, newly matched DM in Fig.~\ref{fig:compare} shows denser 
background reconstruction, but is unable to reconstruct foreground objects properly (the traffic light pole).
By using the predicted and matched DM, we can combine best attributes of both reconstructions.
This way, a denser map with less outliers is generated.



\begin{figure}[!t] 
  \centering
    \includegraphics[width=0.76\textwidth]{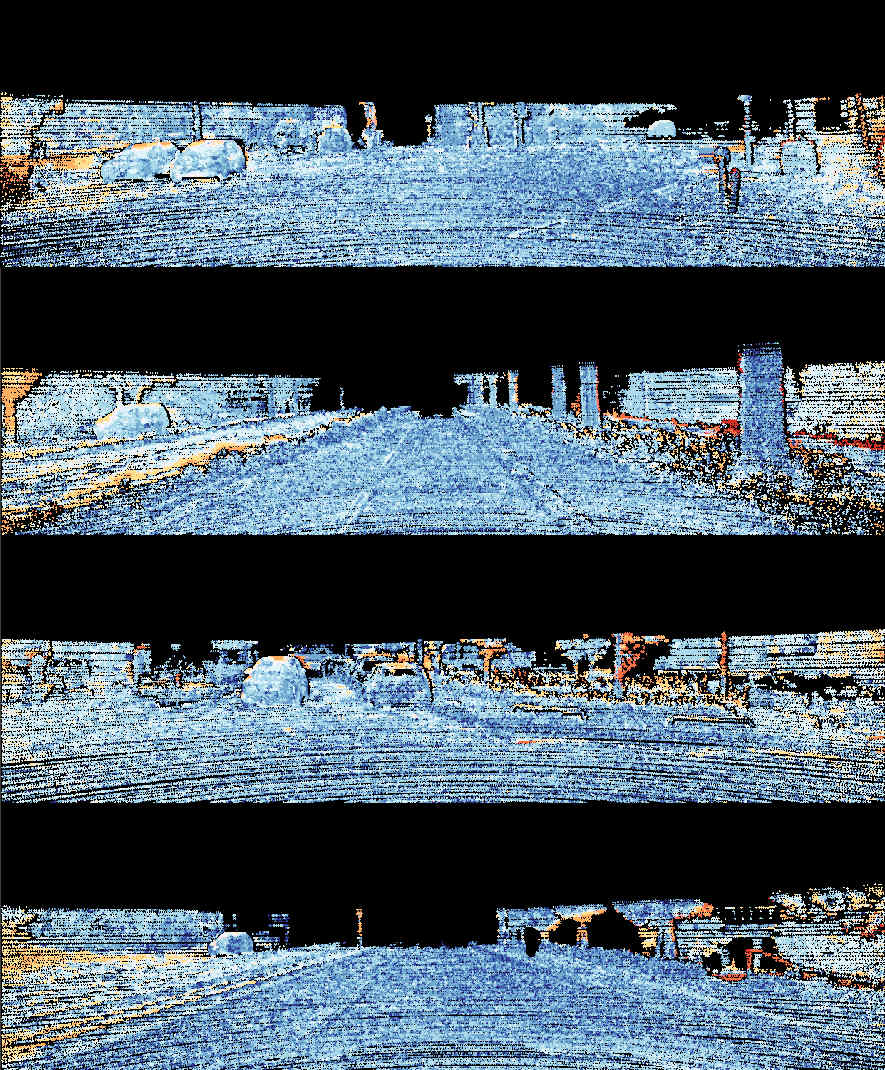} 
    \caption{
    Direct output of KITTI benchmarking script for scene flow training scenes
    130, 84, 87 and 146 respectively (red - bad disparity, blue - good disparity).
    }
    \label{fig:kitti}
\end{figure}

In order to evaluate the proposed approach, we used KITTI benchmarking scripts, where a bad disparity refers to
disparities with error grater than $3$ pixels or $5\%$ relative to the ground truth disparities acquired by the 3D laser
range sensor.
Direct output of the evaluation for four different scenes is shown in Fig.~\ref{fig:kitti}. 
As Table~\ref{t:error} shows, our approach results with less bad disparities than the classical eight-path SGM, while
considering less than 50\% of the whole disparity search space.
This way we increased both computation speed and accuracy of DM estimation. 
Our experiments indicate that this also implies about 50\% reduction of execution time when compared to base SGM.

\begin{table}[!t]
\caption{Comparison of the proposed and basic SGM stereo matching accuracy on the KITTI benchmark scenes}
\centering
	\begin{tabular}{ c | c | c | c } \label{t:error}
    Scene & SGM & Proposed & Proposed, interpolated	\\ \hline
    130.		& 	14.87\% & 	10.19\% 	& 	4.33\%\\ \hline
    84. 		& 	15.55\% & 	11.79\% 	& 	6.34	\% \\ \hline
    87. 		& 	18.13\% & 	12.24\% 	& 	6.51\% \\ \hline
    148.		& 	12.25\% & 	9.75	\% 	& 	5.17	\% 
  	\end{tabular}
\end{table}

\section*{Conclusion and Further work}

In this paper we have presented a stable, accurate, and efficient spatio-temporal disparity estimation algorithm.
The proposed approach is based on using ego-motion between consecutive frames to transform the
DM from the previous frame to the current one.
The transformed disparity, i.e., the predicted disparity, is used as a reference in the disparity search space,
which is reduced based on the predicted disparity uncertainty.
The newly matched disparity from the reduced search space is then used as a measurement within the Kalman filter
 in order to update the predicted disparity.
This results with reduced computational effort and an increase in the accuracy of estimated DM compared to the basic
SGM.
We constrained our use-case to scenes without moving objects, since our primary aim was to achieve accurate
reconstruction of the static parts of the scene, which will then serve as a base for dense scene flow estimation and
moving object detection.
Moreover, the proposed approach can be used in applications with no moving objects, e.g., in space
exploration robotics.
We tested the algorithm on the KITTI benchmark and shown that it can achieve better accuracy than the basic SGM
implementation, while reducing the disparity search space.

\balance


\bibliography{library,fule,additional}             

\end{document}